            [2005/11/29 v0.1.4
 CESCG proceedings sample file]

\documentclass{cescg}[2005/11/12] 
\usepackage{graphicx}

\usepackage{hyperref}

\usepackage{subfigure}

\title{A Client/Server Based Online Environment for Manual Segmentation of Medical Images}

\author{Paper09}
 \author{Daniel Wild\thanks{daniel.wild@student.tugraz.at} %
         \and Maximilian Weber\thanks{m.weber@student.tugraz.at} %
 \and Jan Egger\thanks{egger@icg.tugraz.at}}
 \affiliation{Institute of Computer Graphics and Vision\\
              Graz University of Technology\\
              Graz~/~Austria\\~\\
              Computer Algorithms for Medicine Lab\\
              Graz~/~Austria
              }

\begin{document}

\maketitle

\begin{abstract}
Segmentation is a key step in analyzing and processing medical images.
Due to the low fault tolerance in medical imaging, manual segmentation remains the de facto standard in this domain.
Besides, efforts to automate the segmentation process often rely on large amounts of manually labeled data.
While existing software supporting manual segmentation is rich in features and delivers accurate results, the necessary time to set it up and get comfortable using it can pose a hurdle for the collection of large datasets.

This work introduces a client/server based online environment, referred to as Studierfenster, that can be used to perform manual segmentations directly in a web browser.
The aim of providing this functionality in the form of a web application is to ease the collection of ground truth segmentation datasets. Providing a tool that is quickly accessible and usable on a broad range of devices, offers the potential to accelerate this process.
The manual segmentation workflow of Studierfenster consists of dragging and dropping the input file into the browser window and slice-by-slice outlining the object under consideration. The final segmentation can then be exported as a file storing its contours and as a binary segmentation mask.

In order to evaluate the usability of Studierfenster, a user study was performed.
The user study resulted in a mean of 6.3 out of 7.0 possible points given by users, when asked about their overall impression of the tool.
The evaluation also provides insights into the results achievable with the tool in practice, by presenting two ground truth segmentations performed by physicians.
\end{abstract}


\keywordlist

\section{Introduction}

Image segmentation is an important step in the analysis of medical images. It helps to study the anatomical structure of body parts and is useful for treatment planning and monitoring of diseases over time.
Segmentation can either be done manually by outlining the regions of an image by hand or with the help of semi-automatic or automatic algorithms.
A lot of research focuses on semi-automatic and automatic algorithms, as manual segmentation can be quite tedious and time-consuming work.
But even for the development of automated segmentation systems, some ground truth has to be found, telling the system what exactly constitutes a correct segmentation.
As the avoidance of errors is of especially high importance in the medical domain, this ground truth is typically delivered by physicians, as they have the necessary expert knowledge to reliably decide what should be part of a segmentation and what not.
To form the ground truth physicians can make use of manual or semi-automated segmentation techniques.

Software supporting such techniques usually requires a local installation on the computer of the user and some training time to get comfortable with the extensive features available. This necessary preparation time can be a hurdle for the collection of a big amount of data, which is typically needed to develop a robust automated segmentation algorithm.
Web-based tools have the advantage that they require no installation time and can be adapted quickly to the needs of a particular data collection task, without having to distribute updates to each user individually.

This work describes the development of a client/server based online environment for manual segmentation, referred to as Studierfenster (\url{www.studierfenster.at} or \url{http://studierfenster.tugraz.at}) in the rest of the work, that can be used directly in a web browser. Thus there is no need to install any software, as the tool is readily available in the web browser of the user. Another advantage is the platform independence of Studierfenster, which keeps its potential user base as big as possible.

Following this first outline of the motivation behind the development of Studierfenster, \autoref{relatedwork} briefly discusses related work. Sections~\ref{architecture} and \ref{workflow} then go into more detail on the implementation of Studierfenster. The expert evaluation described in \autoref{expert} and the user study described in \autoref{studyofusers} constitute the evaluation of the tool.
The final~\autoref{conclusion} gives a short conclusion and suggests areas of improvement and approaches for future work.

\section{Related Work} ´\label{relatedwork}
There is a wide range of offline software tools available that offer sophisticated medical image analysis and processing capabilities, including tools for manual segmentation. Examples include 3D Slicer~\cite{slicerGBM} and MeVisLab~\cite{MeVisLabOpenIGT}, ~\cite{MeVisLabVive}. The drawback of these tools is that they require a local installation and are thus not readily available on any device.

Examples of actively supported web-based tools useable for medical imaging include the OHIF Viewer (\url{https://viewer.ohif.org}), Paraview Glance (\url{https://kitware.github.io/paraview-glance}) and Slice:Drop~\cite{slicedrop}, the medical image viewer used as the basis for the development of Studierfenster. While some tools, like the OHIF Viewer, provide simple analysis tools like a ruler for length measurements or annotation tools, others, like Paraview Glance and Slice:Drop, provide only visualization capabilities. What is missing is a tool that provides more sophisticated analysis tools or even processing routines directly in the web-browser. 

\section{Software Architecture}\label{architecture}
This section gives a high-level overview of the architectural model behind the segmentation tool developed in this work.

\begin{figure}[]
\includegraphics[scale=0.34]{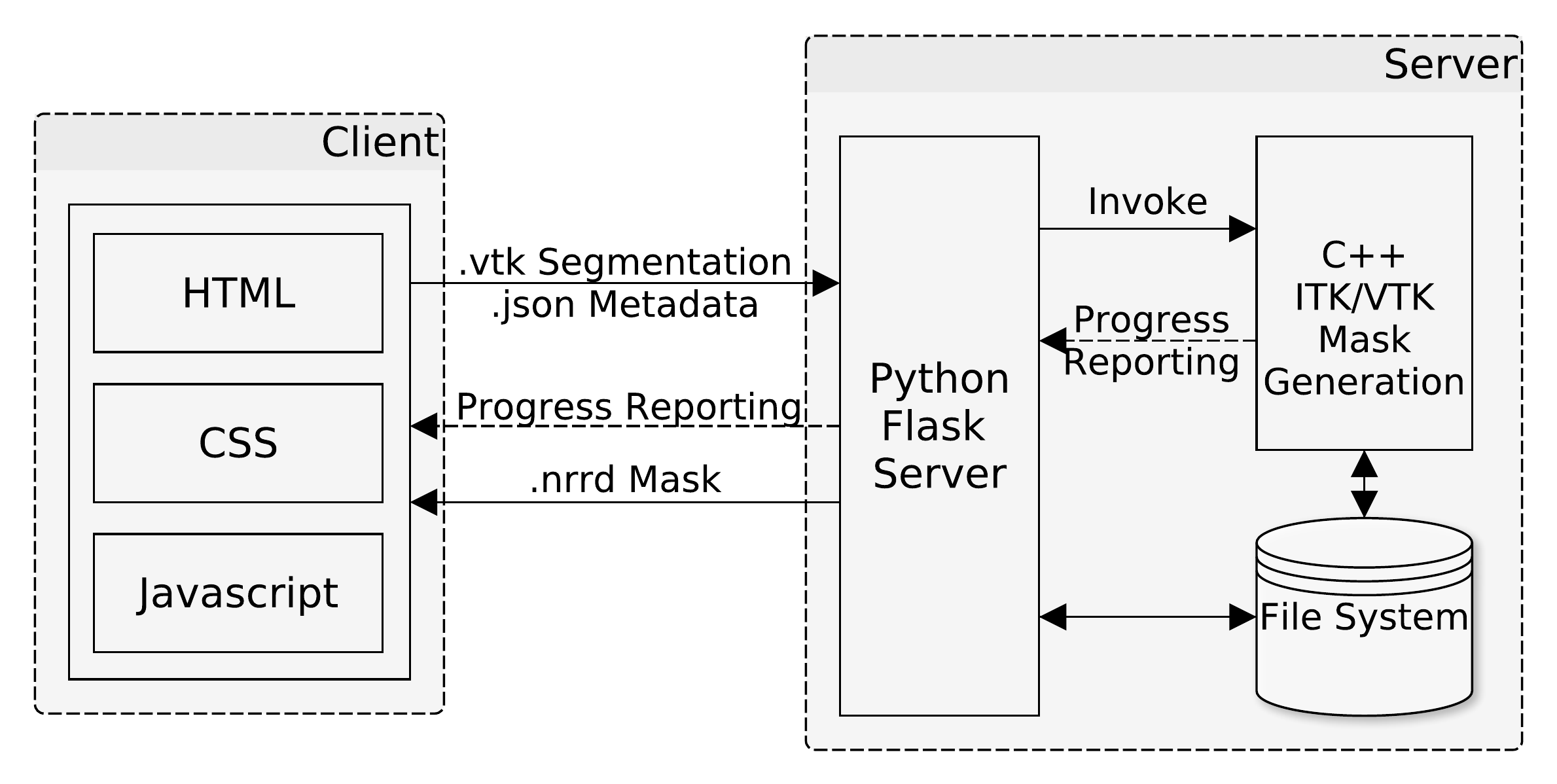}
\caption{An architectural model of the mask generation process.} \label{arch}
\end{figure}

Slice:Drop, which is the medical image viewer this tool is built upon, uses a purely client-oriented approach. This means that all the necessary computations for visualization are carried out on the client using JavaScript and no files have to be uploaded to the host server. The most recent version of Slice:Drop at the time of writing is limited in its functionality, however. While it does provide a volume rendering view and three 2D views in the axial, sagittal and coronal direction, it does not provide any image processing capabilities. For the presented segmentation extension of Slice:Drop, an additional server backend was developed. An exemplary use case of the server backend is the conversion of the segmentation contours to a filled Nearly Raw Raster Data (NRRD) volume mask. The architectural scheme behind this conversion can be seen in~\autoref{arch}. In the following, this conversion process will be used to describe the architecture behind the segmentation tool in more detail.

\subsection{Client}
The process starts on the client side with the user opening up a web browser and navigating to the domain that hosts the HTML, CSS and JavaScript files of the segmentation tool. These files are then transferred to the user so that the segmentation tool can be used directly in the web browser.

Now the user is able to load a volume file in the viewer and carry out a segmentation of some area of interest. The resulting segmentation contours are all stored in JavaScript variables for now. If the user decides to save the segmentation contours directly as a file in the Visualization Toolkit (VTK)~\cite{vtkbook} format and without conversion to an NRRD volume mask, no additional communication with the server is necessary. The process of carrying out a segmentation and saving only its contours as a VTK file thus makes no use of the server backend.

\subsection{Server}
The primary component of the server backend is the Python Flask server instance, as it is responsible for handling all the communication with the client.
The first moment this Python Flask server comes into play is during the process of converting the segmentation contours to an NRRD volume mask.
If the user decides to do so, the segmentation contours are internally written to a VTK file on the client. This file is then immediately uploaded to the server, however. To be able to construct an NRRD volume mask with the same size, spacing, and orientation as the original file, the corresponding information of the original file has to be uploaded additionally. In order to not having to upload the entire original file, this metadata is extracted from the file on the client and then uploaded to the server as a small separate JSON file.

Once the segmentation and the metadata file are both uploaded, the client asks the server to invoke the C\texttt{++} volume mask generation program, which is explained in more detail in~\autoref{cplusplus}. During the conversion process, the C\texttt{++} program continuously reports the progress to the Python Flask server, which then hands over the progress information to the client.

As soon as the conversion is finished, the C\texttt{++} program writes the NRRD volume to the file system of the server, from where it is handed over by the Python Flask server to the client for downloading.

Due to the modular design of the server backend, it is also possible to integrate other processing capabilities into it. An exemplary additional server module, which was integrated to prove said modularity, is a volume file converter. The converter takes volume files as input and produces a volume file in the NRRD file format. Additionally, the volume files are transformed into a world coordinate system with a Right Anterior Superior (RAS) basis. An explanation of this coordinate system is given in~\autoref{contours}.

\section{Segmentation Workflow}\label{workflow}
\begin{figure}[htb!]
\includegraphics[scale=0.38]{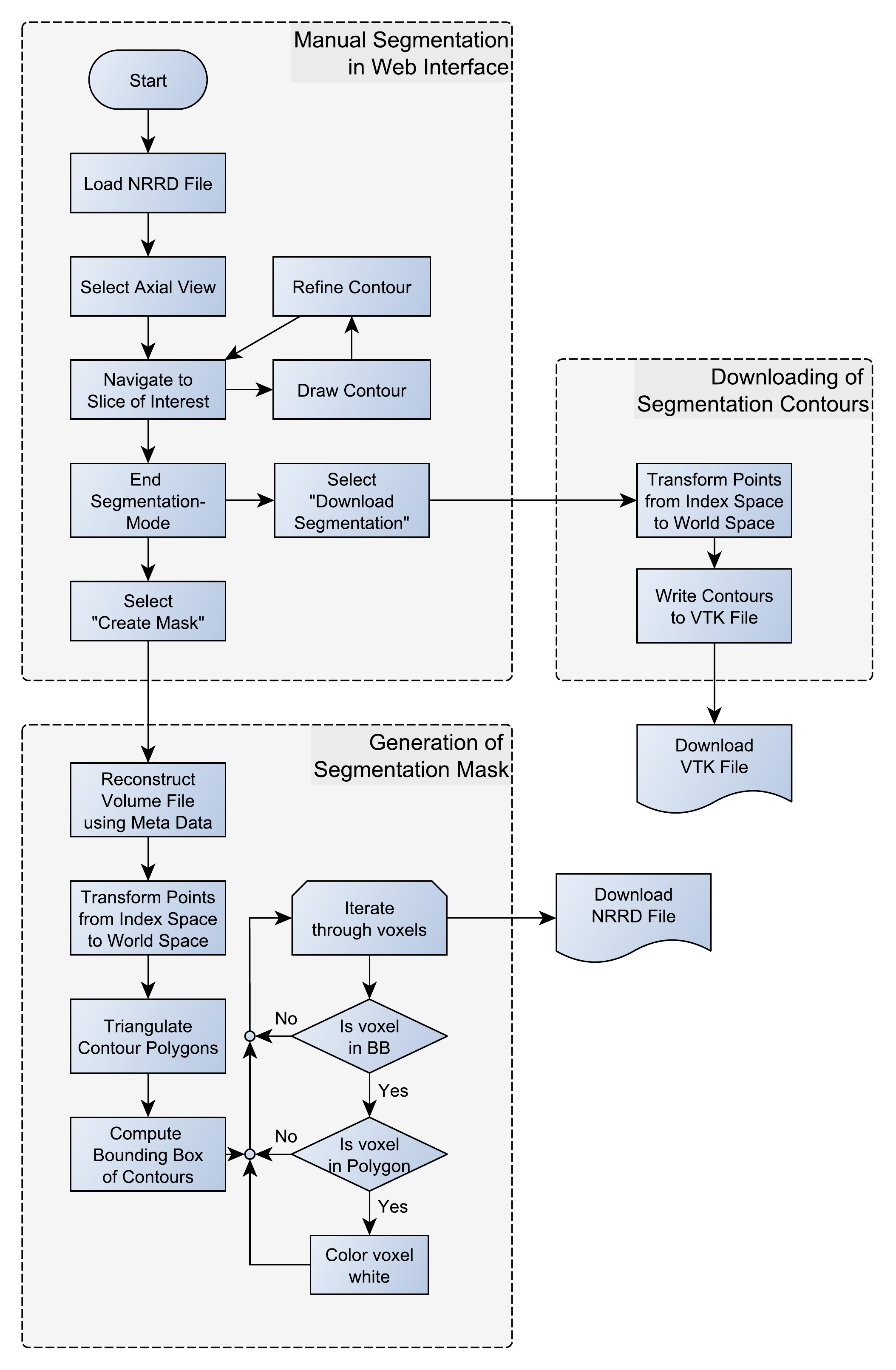}
\caption{This figure shows the workflow to obtain both a VTK file of the segmentation contours and a filled NRRD volume mask.}\label{flowchart}
\end{figure}
The following section will explore the segmentation workflow of the tool developed in this work in more technical detail. The flowchart in~\autoref{flowchart} illustrates the necessary steps to obtain both a mesh file containing the segmentation contours and a volume file containing a filled segmentation mask. The process can be grouped into three main parts: The manual outlining of the object to be segmented in the web interface, the conversion of these outlines to segmentation contours stored in a VTK file and the generation of a filled segmentation mask on the server backend.
\subsection{Manual Segmentation in Web Interface}

\begin{figure}
  \centering
  \includegraphics[scale=0.3]{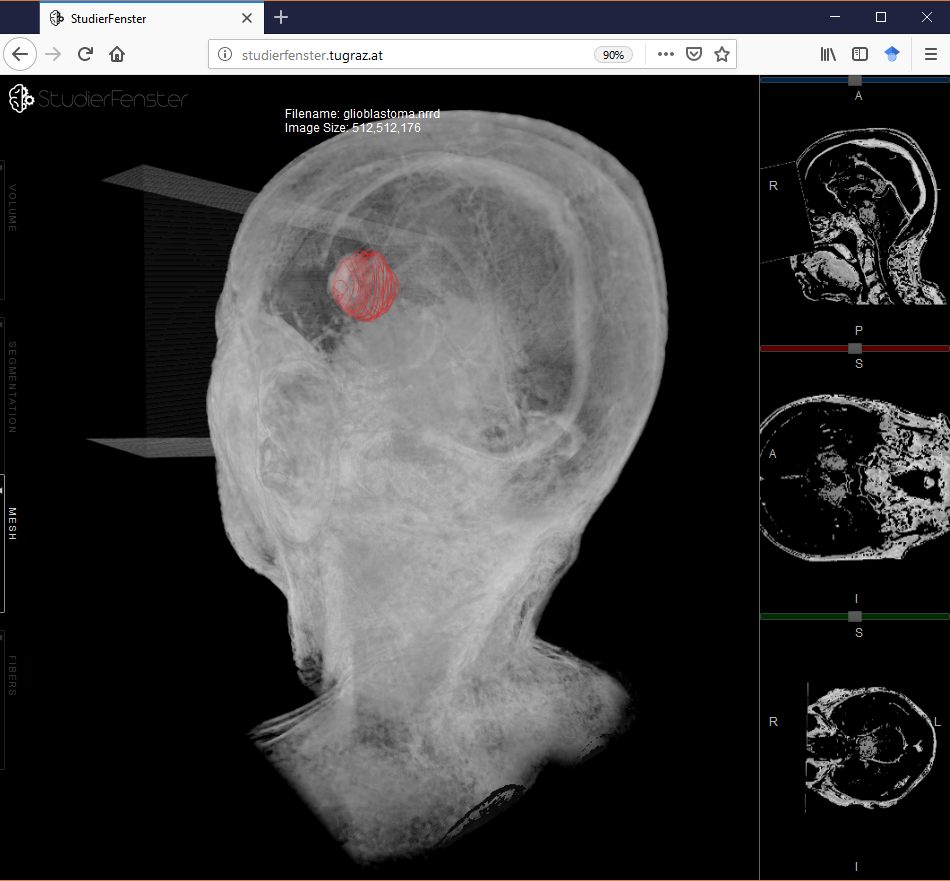}
  \caption{Contours of a glioblastoma segmentation visualized in Studierfenster.}
  \label{interop}
\end{figure}

The interface used for drawing the manual segmentation contours is developed as an extension to the web interface of Slice:Drop, the medical image viewer Studierfenster is based upon.

Slice:Drop's core feature is loading and viewing volume files directly in the web browser. So for the first step in the segmentation workflow, which is loading the NRRD volume file of interest into the segmentation tool, the standard loading method of Slice:Drop can be used. Selecting a file to load can either be done by a simple drag\&drop interaction or by using the file finder of the web browser. Once a file is selected Slice:Drop calls the NRRD file parser of the X Toolkit (XTK)~\cite{slicedrop} to perform the loading of the file. Some changes had to be made to this file parser to allow Slice:Drop to extract the space and orientation information from the header of the NRRD file. This information is needed later on during the mask generation process in order to reconstruct a mask with the same origin, spacing, and size as the original volume file.

As soon as the parsing is complete, the volume is displayed in the predefined views of Slice:Drop. Slice:Drop makes use of the renderer classes of XTK to perform the necessary reslicing and volume ray casting for 2D and 3D visualization respectively. In the default setting the 3D view is enlarged and the three 2D views are miniaturized on the right side.
As the segmentation extension only supports segmenting in the 2D axial view, the user now has to enlarge this view by clicking on it.

On the axial view, one can now navigate through the different slices of the volume file by either using the mouse wheel or the slider on top of the view. Once the first slice featuring an element of interest is in view, the segmentation mode can be started by selecting ``Start Segmentation" in the segmentation menu on the left side. In order to display the contours, a second HTML canvas with a transparent background was superimposed on the canvas displaying the current slice. As the HTML canvas itself does not provide any drawing capabilities, the standard drawing methods of its \texttt{getContext("2D")} object are used to display the points of a segmentation and its connecting lines on the drawing canvas. Entering the segmentation mode brings the drawing canvas into the foreground.

The user can now start to draw a segmentation contour by holding the left mouse button pressed while moving along the border of the object of interest. A new point is added to the segmentation contour once the current distance of the mouse cursor to the previous point surpasses a given threshold. The segmentation contour can be finished by moving the mouse cursor close to the first point of the contour and releasing the left mouse button. To contrast the newly segmented region from the rest of the slice, the segmented region is filled with a light red color.

If the user is not completely satisfied with the accuracy of the contour, individual points can be deleted and reset to a new location. This can be done by clicking on the erroneous points and subsequently clicking on the correct location. Should the user wish to start over with a segmentation contour completely, the whole contour can be deleted by selecting ``Delete Slice" and clicking on a point lying on the contour to be deleted.

As soon as every region of interest on one slice is segmented with satisfying accuracy, the user can navigate to the next slice using the mouse wheel and start over with the process of drawing and refining the segmentation contours.
Once the segmentation on all individual slices containing the object of interest is finished, the user can download the combined contours directly as a VTK file or convert them to a filled volume mask in the NRRD format. The technical details behind those two options are explained in the following two sections.

\subsection{Download of Segmentation Contours}\label{contours}

The quickest way to export the contours of a segmentation is to download them as a VTK file. As explained in~\autoref{architecture}, this can be done without using the server backend. The contours of the segmentation are written to a VTK file directly on the client. Before starting to write contours to the VTK file, however, one has to take care to align the coordinates of their points with the coordinate system of the segmented volume file.

During the segmentation process, a point is internally stored in index space coordinates. Described informally, in index space the coordinates of a point directly refer to indices into the volume file. Indexing into the volume file with the coordinates of a point would thus yield the voxel at which the point is located.
The volume file itself, however, is stored in a coordinate system called world space. This coordinate system describes the position and orientation of a patient relative to the medical scanner used to acquire the image volume file. It is defined by the origin, which is the position of the first voxel of the volume file in millimeters and its basis vectors. A commonly used basis in neuroimaging, which is also used by Slice:Drop as the reference frame to display volume files in, is the RAS basis. Its axes are related to the patient being scanned, with the R axis increasing from left to right, the A axis increasing from posterior to anterior and the S axis increasing from inferior to superior~\cite{coordinates}.

In order to align the segmentation contours with the volume file, the points of the contours now have to be positioned in the world space as well. This is done by multiplying each point with the \texttt{IJKToRAS} Matrix provided by XTK, which describes an affine transformation from the index coordinate system to the RAS coordinate system.

Once the transformation of a point is completed, its coordinates are written to the first part of the VTK file. While the first part of the file comprises the coordinates of all points in the segmentation, it gives no information on which contour they belong to yet. The connectivity information is defined by the second part of the VTK file. Each line in this part describes one closed contour, by listing the indices of the points in the first part of the file in the order they are connected with one another~\cite{vtkguide}.

\autoref{interop} shows exemplary segmentation contours which were exported as a VTK file and visualized in Studierfenster.

\subsection{Generation of Segmentation Mask}\label{cplusplus}
The second way to export the segmentation is as a filled volume mask in the NRRD format. Such a mask has the same dimension and orientation as the original volume file which was segmented and can thus be superimposed on it. In the volume mask, the voxels lying inside the segmentation contours are white and the voxels outside the contour are black.
Thus the goal is to convert the segmentation contours to such a filled volume mask. This process is handled by a C\texttt{++} program on the server backend, that makes use of both the Insight Toolkit (ITK) and the VTK library. The details of how the client communicates with the server and transmits the necessary files for the conversion are given in~\autoref{architecture}.
Assuming that the two necessary files for the conversion, namely the VTK file containing the segmentation contours and the JSON file containing the space metadata, are present on the server, the conversion process can be started.

The first step is loading the JSON file containing the space metadata and, using said metadata, constructing an all black volume with the same dimensions, origin, and spacing as the original volume file.
For this, the \texttt{Image} class of ITK is used.

The next thing to take care of is the loading of the segmentation contours. As those were transmitted to the server as a VTK file containing polygons, the easiest way to load them is to use the \texttt{vtkPolyDataReader}.
The polygons are given in index space coordinates, however, and thus do not align with the just created volume.
So before continuing the points of the polygons have to be transformed into the same world coordinate system the volume uses. This is done by applying the \texttt{TransformContinuousIndexToPhysicalPoint} method of the ITK Image object the volume is stored as to each individual point of the polygons.

The next, at the first glance seemingly unnecessary step, is to triangulate the polygons using the \texttt{vtkTriangleFilter}. This is done because later on the \texttt{PointInPolygon} method of the \texttt{vtkPolygon} objects is used to check whether a voxel lies within a contour. During development this method yielded bad results with more complex contour shapes if they were used as one big polygon. Splitting the contours up in individual triangles results in the method working reliably, however.

Before now starting to iterate through the volume and checking whether every voxel on every slice lies within one of the polygons, the bounding box of the segmentation is calculated.
Checking if a voxel lies within the bounding box is less time consuming than checking whether it lies within one of the polygons and so incorporating this prior check results in a considerable performance increase.
Only if a voxel lies within the bounding box, an inclusion test with the polygons lying on the voxels slice is performed. This is done with the \texttt{PointInPolygon} method of the \texttt{vtkPolygon} class.
Every voxel that lies within one of the polygons is colored white. Once all the voxels have been iterated through, the resulting volume is thus a binary mask, that marks the area inside the segmented region with white colored voxels.
This volume mask can now be downloaded in the NRRD file format.

\section{Expert Evaluation} \label{expert}
In order to test the segmentation capabilities of Studierfenster, two manual segmentations have been performed by physicians. The resulting segmentation files are then checked for their validity and compatibility with other medical imaging platforms.
\subsection{Datasets} \label{datasets-expert}
The first segmentation was done on an expansive basalioma of the left midface, found in the initial Magnetic Resonance Imaging (MRI) scan of the patient suffering from it. The tumor showed intra orbital growth and was initially unresectable due to its large size. After administration of medicines, the tumor shrank in its size, which eventually allowed surgeons to completely remove it.

The second segmentation was performed on the MRI of a female, 75-year-old patient with a glioblastoma in the left hemisphere. 
\subsection{Results}
\begin{figure}[]
\centering
\includegraphics[scale=0.15]{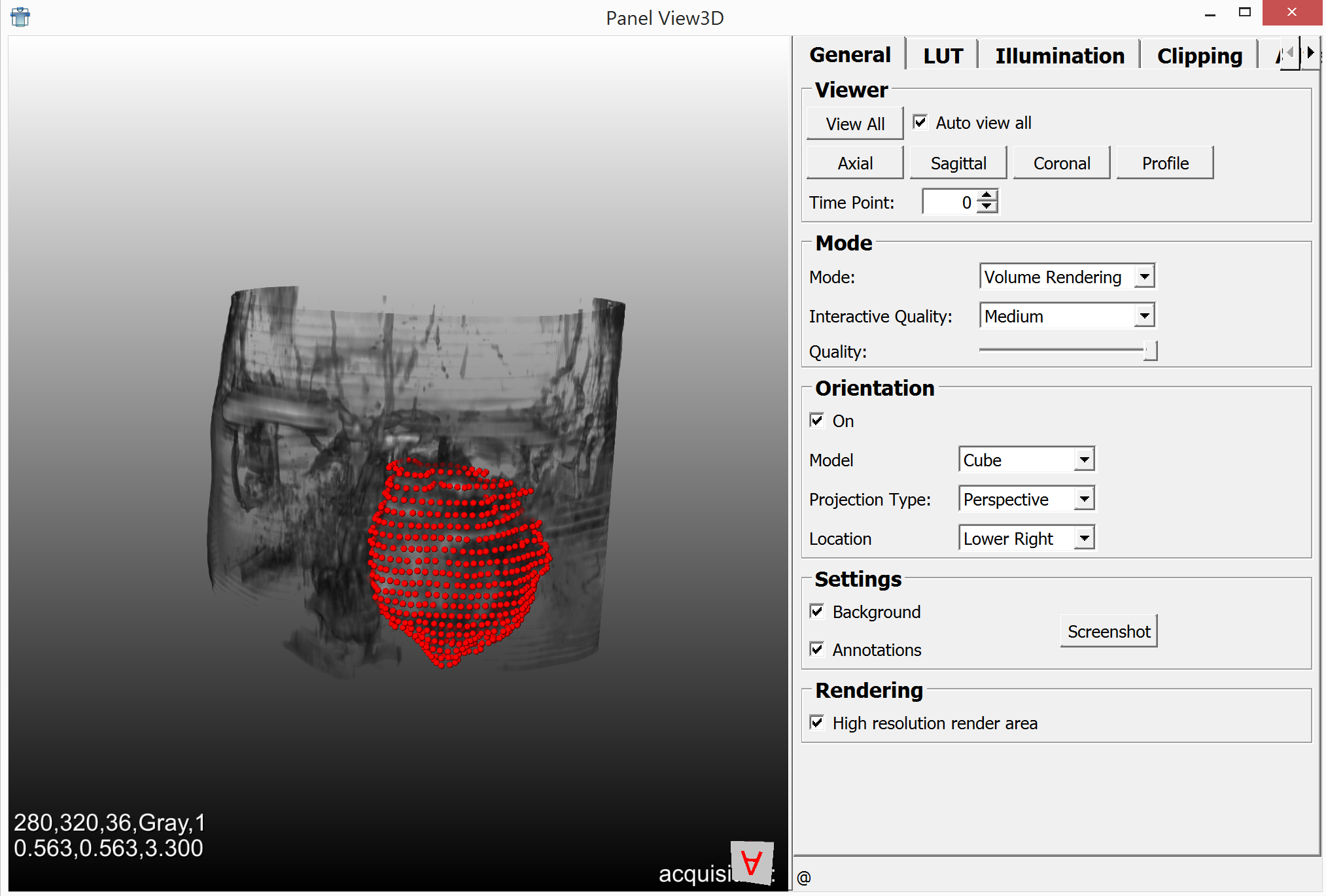}
\caption{Basalioma segmentation contours visualized in MeVisLab. The red point cloud represents the resulting segmentation contours of the basalioma superimposed on the original MRI dataset.} \label{basalmevis}
\end{figure}

The resulting contours of the basalioma segmentation, which took 35 minutes to perform, can be seen in~\autoref{basalmevis}. The visualization is done using the \texttt{View3D} module of MeVisLab, which can be used to superimpose the contours on the original MRI dataset.

\autoref{interop} shows the resulting segmentation contours of the second performed segmentation, which was the one of the glioblastoma, visualized in Studierfenster.

\subsection{Discussion}
The discussion of the obtained segmentation results focuses on two main aspects: The accuracy of the resulting segmentation contours and the compatibility of the file format used to store them with different medical imaging platforms. Both aspects are important prerequisites for the practical application of the tool. To stress their importance one can take the use case of ground truth data collection as an example. Here, wrongly aligned or calculated coordinates of the segmentation contours would render the obtained datasets worthless and annihilate all the effort that has gone into collecting them. Therefore it is important that the segmentation contours in the exported VTK file are at the exact location the expert performing the segmentation intended them to be.
The aspect of file compatibility also comes into play when thinking about the data collection use case. To make the most use of the resulting data, it should be possible to analyze and process the resulting files with already established software without much effort.

One empirical way to verify the two aforementioned aspects is to use a different medical imaging software like MeVisLab to superimpose the exported segmentation contours on the datasets they originate from. That way one can check whether the calculation of the coordinates of the points constituting the segmentation contours has been done correctly and whether the coordinate system of the contours matches the one of the original dataset. This verification was done using the two collected expert segmentations, namely the one of the basalioma seen in~\autoref{basalmevis} and the one of the glioblastoma seen in~\autoref{interop}.
From the screenshots taken in MeVisLab and Studierfenster, it is evident that both expert segmentations are correctly aligned with the datasets they were performed on.
Under the assumption that MeVisLab handles the VTK format correctly, this also demonstrates that Studierfenster exports the contours as valid VTK files, that can be loaded in different medical imaging tools as well.

\section{User Study} \label{studyofusers}
The goal of this work is to develop a browser-based manual segmentation tool. From a user perspective, major usability improvements of a browser-based solution compared to a desktop solution include its faster accessibility due to the missing installation and update process. This advantage is, of course, neglectable if the rest of the tool is not perceived well in terms of usability. Thus in order to evaluate the usability of the presented segmentation tool, a user study was conducted.
\subsection{Dataset} \label{datasets-userstudy}
The ground truth reference for the user study is the expert segmentation of the glioblastoma described in more detail in \autoref{datasets-expert}. It serves as the reference to which the segmentations of the participants of the user study are compared to.

A separate dataset~\cite{dataset} was used during the introduction of Studierfenster that the participants received.
The dataset used for this purpose originates from a clinical evaluation of segmentation algorithms~\cite{mandbone}, ~\cite{ctdata}. It includes ten Computed Tomography (CT) images in the NRRD format of patients without teeth, which were randomly chosen from a bigger dataset.

\begin{figure}[!htp]
\centering
\includegraphics[scale=0.45]{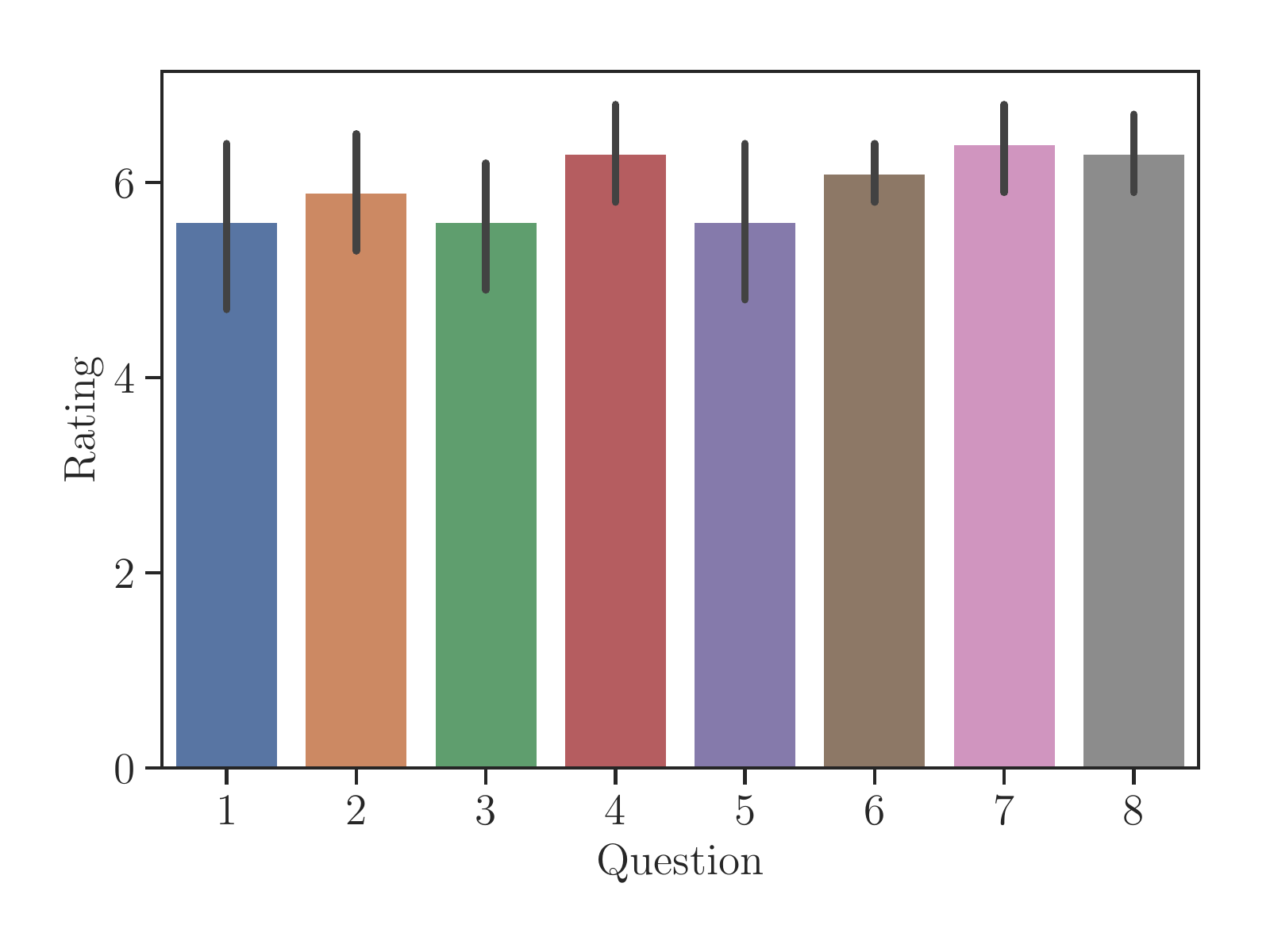}
\caption{Result of the user study visualized as a bar chart. The bars represent the mean of the ratings of all users grouped per question.} \label{barplot}
\end{figure}

\begin{figure*}[!htp]
  \centering
  \subfigure{\includegraphics[scale=0.33]{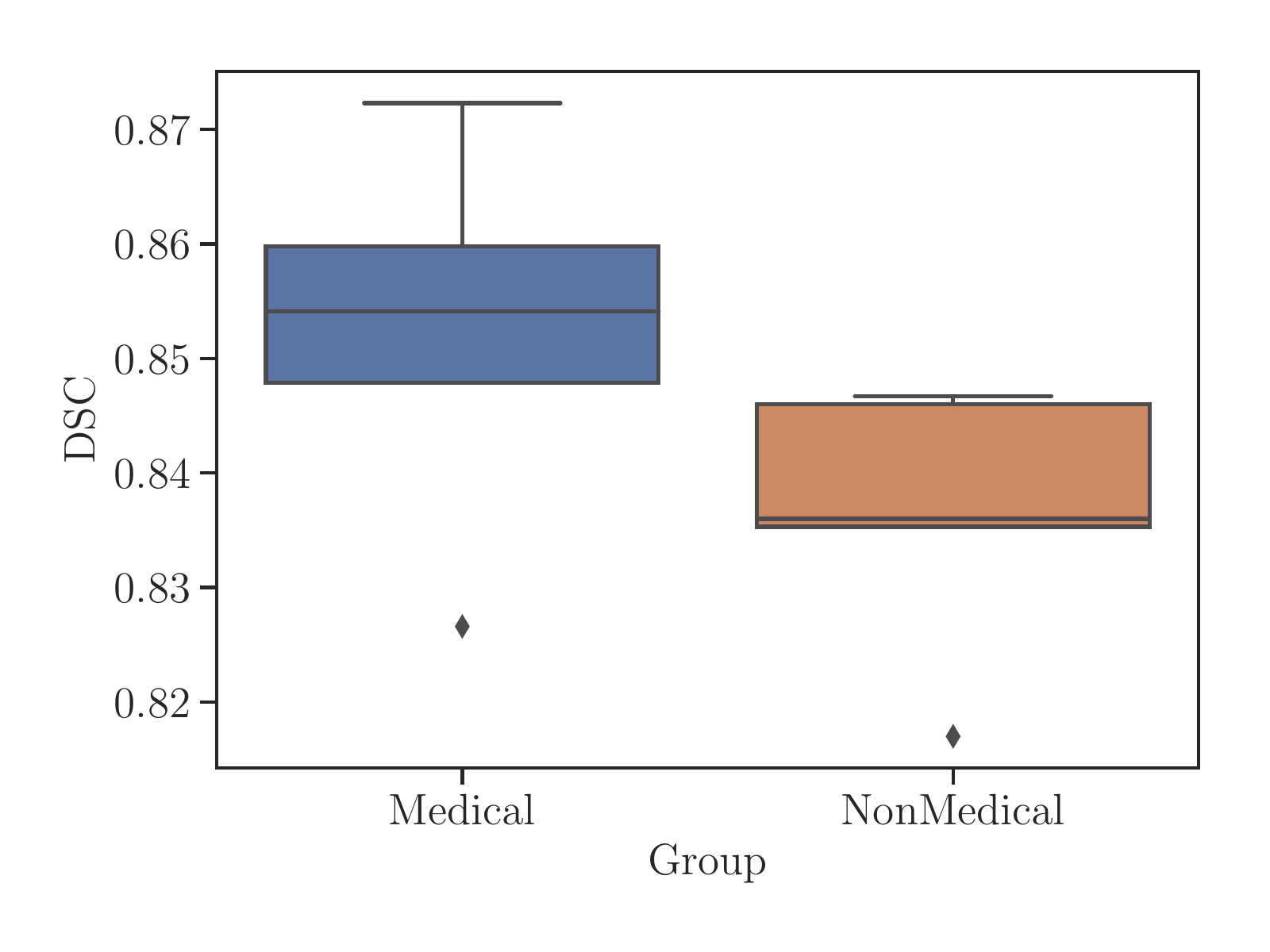}}\quad
  \subfigure{\includegraphics[scale=0.33]{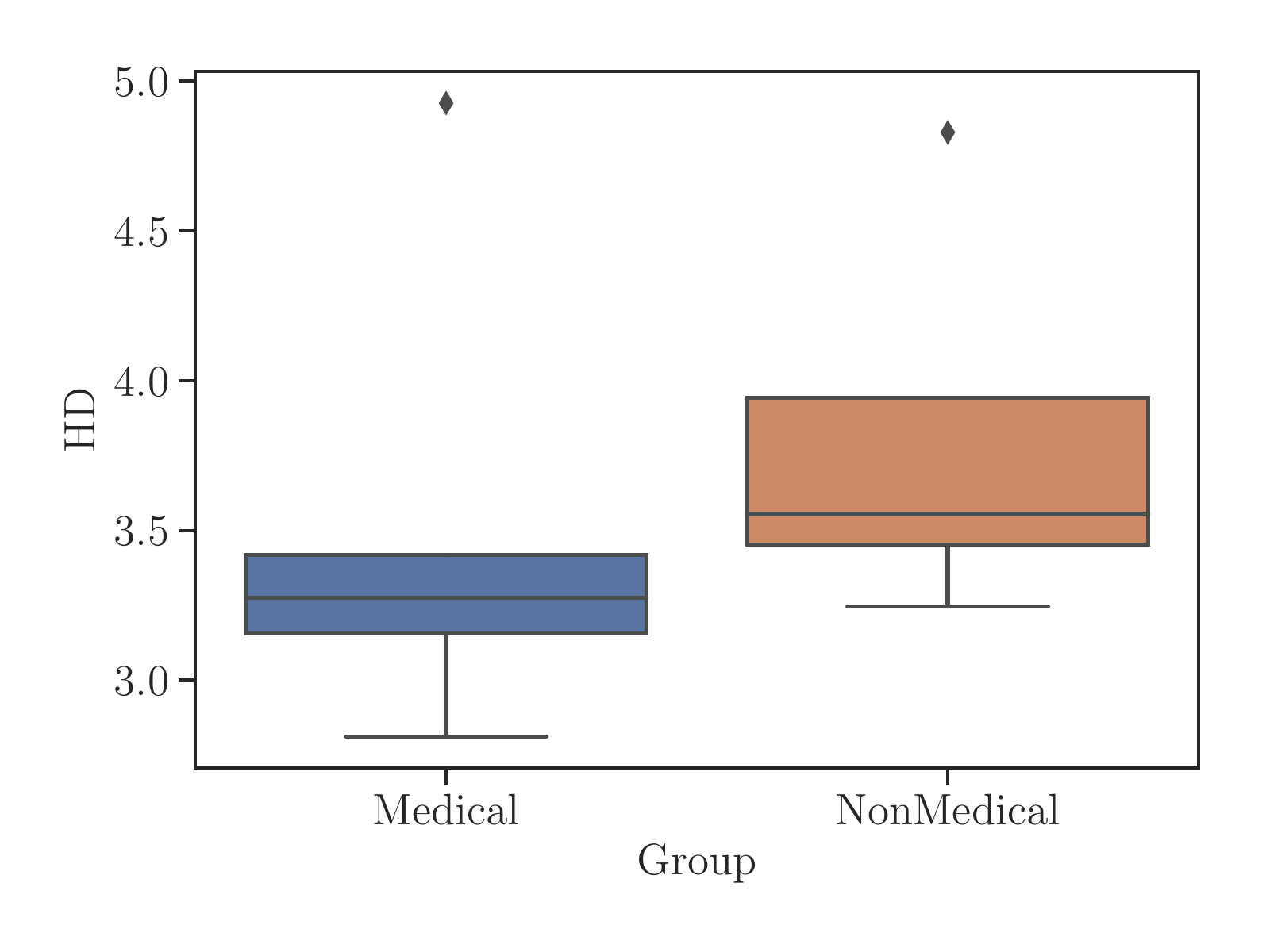}}\quad
  \subfigure{\includegraphics[scale=0.33]{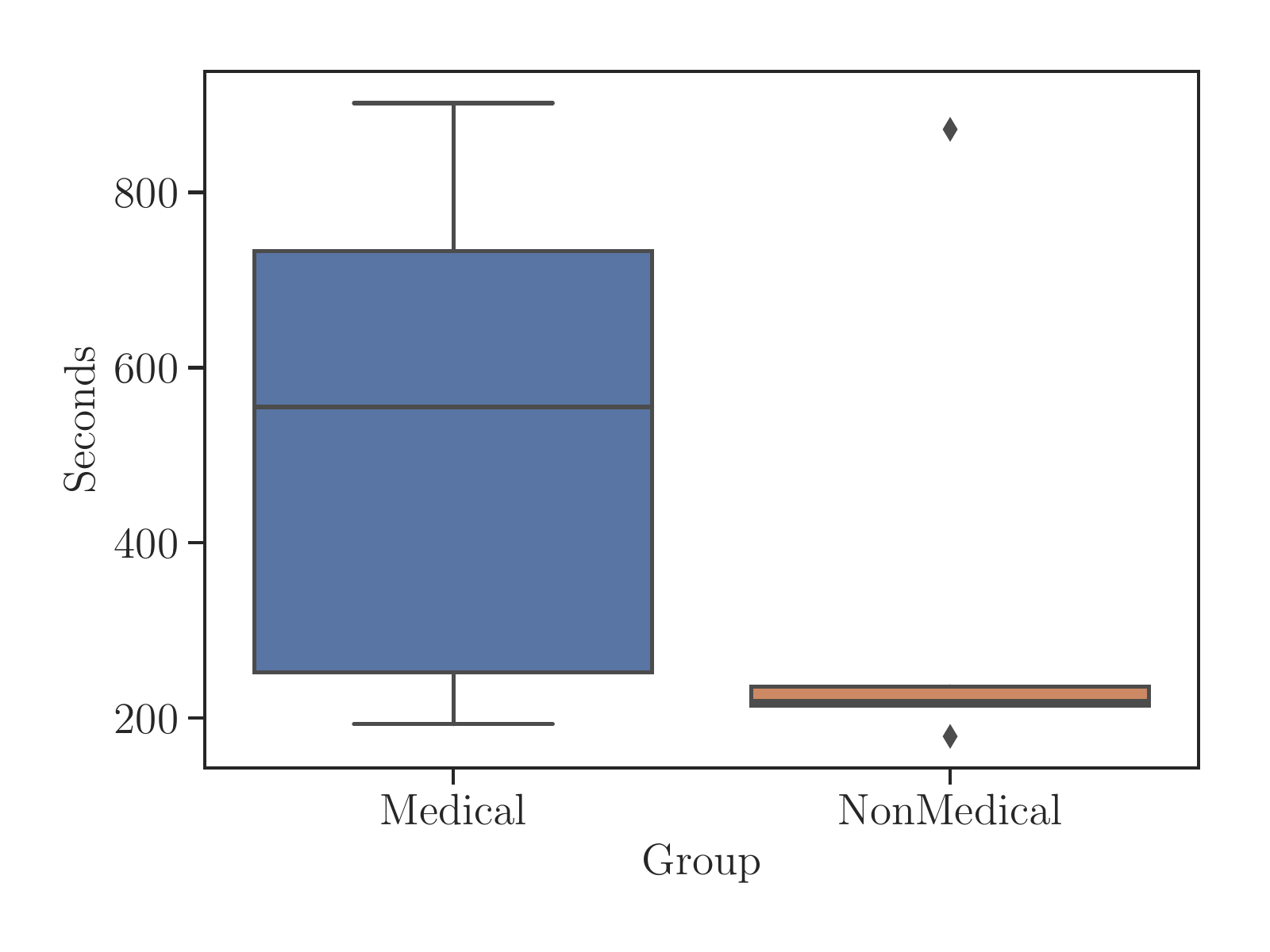}}
\caption{Shown are, from left to right, a comparison of the Dice scores, Hausdorff distances and the time spent on the segmentation task by users with medical background and users with no medical background.} \label{boxplots}
\end{figure*}

\subsection{Methodology} \label{methodology-userstudy}
The design of the user study was derived from the one performed in the context of a cranial implant planning tool for MeVisLab in the work by Egger et al.~\cite{userstudy}. In our case, the test users were first given a short initial introduction to the segmentation tool during which they could explore the features of the platform freely. For this, the CT scan of \emph{patient six} of the mandibular dataset described in~\autoref{datasets-userstudy} was used. For participants with no medical background, the introduction included an explanation of the purpose of the tool and the importance of segmentation for medical imaging. All participants were then guided through the segmentation and metric calculation process.

The actual task for the user was to then segment the same glioblastoma as the one in the reference segmentation seen in \autoref{interop}. This allowed the user to later compare the resulting segmentation with the one of the neurosurgeon. The two comparison metrics, namely the Dice Score~\cite{dice} and Hausdorff Distance~\cite{hausdorff}, were obtained with the help of the calculation tool that was developed as a second use case for the Studierfenster platform by a colleague in parallel to this work. Calculating these two metrics and saving them as a PDF file was the last step of the user's task.
After the user finished the task, a questionnaire was used to capture the impressions of the user regarding the usability of the segmentation tool.
The questions were taken from the work by Egger et al.~\cite{userstudy}, where questions derived from the ISONORM 9241/10 were used. Answers were given on a Likert scale ranging from 1 to 7, where 1 is the worst rating and 7 the best. The questions presented to the user were as follows:
\begin{enumerate}
\item The software does not need a lot of training time.
\item The software is adjusted well to achieve a satisfying result.
\item The software provides all necessary functions to achieve the goal.
\item The software is not complicated to use.
\item How satisfied are you with the UI surface?
\item How satisfied are you with the presented result?
\item How satisfied have you been with the time consumed?
\item How is your Overall impression?
\end{enumerate}

The test users chosen for the user study consisted of 10 users in total. This group was further split into 5 users that are familiar with the medical use of segmentation and the status quo of existing solutions and 5 users that had no prior experience with medical image segmentation.

\subsection{Results}\label{results}
\autoref{barplot} visualizes the mean ratings given in the questionnaire and the corresponding standard error as a bar chart.

In~\autoref{boxplots} the Dice scores of users with a medical background can be seen in comparison with the Dice scores of users with no medical background. \autoref{boxplots} also shows a comparison of the Hausdorff distances.


\subsection{Discussion}
The first part of the user study, which was the short introduction to the tool, with a chance for participants to explore the tool freely, was well received. On average this initial training took about five minutes.

For the actual segmentation task no timing constraints were given to participants. As can be seen in~\autoref{boxplots} the time participants took to finish the task varied between 3 and 15 minutes. It is notable that most of the variance stems from the medical group, which took an average of 8 minutes and 47 seconds, with participants well distributed between the 3 and 15 minutes mark. In contrary all participants but one of the non-medical group took between 3 and 4 minutes for the segmentation task. The one visible outlier took 14 minutes and 32 seconds.
The most probable cause for the time difference between the two groups is that participants of the medical group edited their segmentation contours more frequently.

Once the participants were content with the quality of their segmentation, they were asked to calculate the Hausdorff Distance and the Dice Score between their segmentation and the reference segmentation of the physician. The resulting Hausdorff Distances and the Dice Scores can be seen in~\autoref{boxplots}. Hausdorff Distances range between $2.81$ and $4.93$ and Dice Scores between $0.82$ and $0.87$.
A comparison between the medical and the non-medical group again reveals differences in values, albeit them being less apparent than those found in the segmentation times. The difference between the two mean Hausdorff distances per group is $0.287$, with the medical group having the lower mean distance of $3.518$ compared to the mean distance of $3.805$ found in the non-medical group.
Dice Scores also show slightly better results for the medical group. Here the mean of the medical group was $0.852$ and the mean of the non-medical group $0.836$, yielding a small difference of $0.016$.
Surprisingly the non-medical group did thus not perform much worse than the medical group, even though they took less time to complete the manual segmentation of the glioblastoma.

In the last part of the user study, participants were handed the usability questionnaire described in \autoref{methodology-userstudy}.
As can be seen in the bar chart in~\autoref{barplot}, the ratings were overall positive, ranging between a minimum mean rating of $5.6$ given for questions 1, 3 and 5 and a maximum mean rating of $6.4$ for question 7. Question 1 was about the necessary training time for the tool and question 5 about the satisfaction with the UI surface. Putting this results in the context of the segmentation interface suggests that there is still work to be done to make it more intuitive to use. One specific area to improve would be the presentation of the buttons on the left-hand side. Adding icons to them would improve their look and giving additional information on their function when hovering over them would reduce the necessary training time for the tool. The mean rating of $6.3$ given to question 4 still suggests that the tool is not too complicated to use.
Question 3 asked participants whether the tool provided all necessary functions to achieve the manual segmentation and the calculation of the results. One suggestion regarding an additional feature that was repeatedly given over the course of the user study was the integration of a zoom function. This would ease the segmentation of small structures and in turn improve the segmentation results. The next highest result of the questionnaire was the mean rating of $5.9$ given to the second question asking whether the tool is well adjusted to achieve a satisfying result. Although this rating is already high, it could for example also be improved by including the aforementioned zoom function. However, as the high mean rating of $6.1$ given to question 6 indicates, the tool already allows users to produce satisfying results for sufficiently big structures.

The on average highest rated question shows that participants were content with the time consumed to complete the task and, with a mean rating of $6.3$, the overall impression was also reported as very good.

\section{Conclusion and Future Outlook}\label{conclusion}
This work presented the implementation of a web-based tool supporting manual segmentation directly in the browser. The motivation behind focusing on web technologies was to ease the collection of ground truth segmentation datasets, by providing an easily accessible tool to create them. Considering the positive feedback from the user study and the evaluation of the expert segmentations acquired with the tool, it can be concluded that this goal was reached.
Exported segmentation contours correctly align with their dataset of origin and are compatible with established medical imaging tools such as MeVisLab.

Nevertheless, there remain areas to improve upon in future work, like including a zoom function and other features enabling more advanced segmentation tasks.
One could also think about ways to eliminate the need to upload files to the web server in order to analyze and process them. At the time of writing, this is necessary for generating the segmentation masks and for calculating the segmentation metrics. A promising solution to this could be JavaScript enabled versions of ITK and VTK, which are, at the time of writing, under heavy development by Kitware, the company behind the two libraries, but not yet finalized and completely documented. Including these libraries and the aforementioned features will be considered in the ongoing efforts to expand and improve the presented tool.

\section{Acknowledgments}
This work received funding from the Austrian Science Fund (FWF) KLI 678-B31: ``enFaced: Virtual and Augmented Reality Training and Navigation Module for 3D-Printed Facial Defect Reconstructions'' and the TU Graz Lead Project (Mechanics, Modeling and Simulation of Aortic Dissection). Moreover, this work was supported by CAMed (COMET K-Project 871132) which is funded by the Austrian Federal Ministry of Transport, Innovation and Technology (BMVIT) and the Austrian Federal Ministry for Digital and Economic Affairs (BMDW) and the Styrian Business Promotion Agency (SFG).

\bibliographystyle{plain}
\bibliography{cescg_09}

\end{document}